\documentclass[conference]{IEEEtran}
\IEEEoverridecommandlockouts
\usepackage{cite}
\usepackage{amsmath,amssymb,amsfonts}
\usepackage{algorithmic}
\usepackage{graphicx}
\usepackage{textcomp}
\usepackage{xcolor}
\def\BibTeX{{\rm B\kern-.05em{\sc i\kern-.025em b}\kern-.08em
    T\kern-.1667em\lower.7ex\hbox{E}\kern-.125emX}}
\usepackage{orcidlink}
\usepackage{marvosym}
\usepackage{array}
\usepackage{threeparttable}
\usepackage{multirow}
\usepackage{amsmath}
\usepackage{hyperref}

\begin{document}

\title{Efficient End-to-End 6-Dof Grasp Detection Framework for Edge Devices with Hierarchical Heatmaps and Feature Propagation}

\author{\IEEEauthorblockN{Kaiqin Yang$^{\dagger {1}}$, Yixiang Dai$^{\dagger {1}}$, Guijin Wang$^{1}$, Siang Chen$^{\ddagger {1}}$}
\IEEEauthorblockA{$^{1}$Department of Electronic Engineering, Tsinghua University, Beijing 100084, China.\\
$^\dagger$Equal Contribution.\quad\quad $^\ddagger$Corresponding Author: csa21@mails.tsinghua.edu.cn}
}
\maketitle

\begin{abstract}
6-DoF grasp detection is important for the advancement of intelligent embodied systems, as it provides feasible robot poses for object grasping. Various methods have been proposed to detect 6-DoF grasps through the extraction of 3D geometric features from RGBD or point cloud data. However, most of these approaches encounter challenges during real robot deployment due to their significant computational demands, which can be particularly problematic for mobile robot platforms, especially those reliant on edge computing devices. This paper presents an \underline{E}fficient \underline{E}nd-to-\underline{E}nd \underline{G}rasp Detection \underline{Net}work (E3GNet) for 6-DoF grasp detection utilizing hierarchical heatmap representations. E3GNet effectively identifies high-quality and diverse grasps in cluttered real-world environments. Benefiting from our end-to-end methodology and efficient network design, our approach surpasses previous methods in model inference efficiency and achieves real-time 6-Dof grasp detection on edge devices. Furthermore, real-world experiments validate the effectiveness of our method, achieving a satisfactory 94\% object grasping success rate. More details can be found on our \href{https://kaiqinyang.github.io/E3GNet.github.io}{project page}.
\end{abstract}


\section{Introduction}

Visual-guided robotic grasping is fundamental in embodied intelligence. Despite the significant progress made in single object grasping, it remains an challenge to grasp cluttered objects efficiently and precisely in unstructured environments.

Traditional grasping methods utilized model-based prior with full knowledge of objects' physical models \cite{bohg2013data}, which is challenging to use in real world. Advances in deep learning have driven the development of model-free grasp detection methods based on visual information in open-world scenes. Earlier methods generate 4-DoF planar grasps from a single RGBD observation \cite{kumra2020antipodal,morrison2018closing}. Despite its efficiency, these representations constrain the gripper perpendicular to the camera plane, which sacrifices the degree of freedom and limits the performance in complex scenes. 

Recent advancements in 6-DoF grasp detection have garnered significant attention due to their wide-ranging applications, allowing robots to grasp objects from various orientations. Previous research has proposed methods based on sampling and evaluation to generate 6-DoF grasp poses by sampling and assessing grasp candidates from point clouds \cite{liang2019pointnetgpd, ten2017grasp}. Other approaches have employed deep neural networks to directly regress relevant attributes in an end-to-end manner \cite{ni2020pointnet++}. However, both methods could be more efficient and accurate. More recent studies focus on identifying high graspable regions by incorporating global information from scene point clouds or RGBD images \cite{wang2021graspness, zhao2021regnet, chen2023hggd}. Though effective, these methods primarily utilize 3D feature encoders, which entail higher computational costs and may not be suitable for edge-devices deployment.


In this work, we propose a novel \textbf{E}fficient \textbf{E}nd-to-\textbf{E}nd 6-Dof \textbf{G}rasp detection \textbf{Net}work (\textbf{E3GNet}) based on hierarchical heatmap representations, which succeed in detecting high-quality and diverse grasps for real-world object clutters. To the best of our knowledge, our method is the first to achieve real-time 6-Dof grasp detection on edge devices. Our framework first constructs a Global Location Heatmap FPN (feature pyramid network) with a lightweight encoder, Geometry-aware MobileOne \cite{vasu2023mobileone}, to obtain multi-scale features and predict grasp location heatmaps. Then, the Region Feature Propagation module aggregates graspable region features under the guidance of the location heatmaps. Finally, the graspable region features are fed into a specially designed Rotation Heatmap generation model for grasp rotation detection and refinement, forming scene-level 6-Dof grasps. Experiments on the large-scale grasp dataset show that E3GNet detects more precise and diverse grasps than previous methods on the benchmark. Model inference efficiency experiments on multiple platforms, including edge devices, further prove the computation efficiency of our method. Real-world grasp experiments on the real robot also yield a satisfactory 94\% grasp success rate.

The detailed contributions of this work are as follows.
\begin{itemize}
    \item We propose a novel efficient end-to-end 6-Dof grasp detection framework (\textbf{E3GNet}), realizing real-time 6-Dof grasp detection on edge devices.
    \item We design a novel Region Feature Propagation module and a Rotation-Heatmap-Based Grasp Detection technique to achieve efficient and precise grasp detection.
    \item We develop a Global Location Heatmap FPN combined with a lightweight encoder, Geometry-aware MobileOne to efficiently obtain multi-scale features and locate grasps.
\end{itemize}

\section{Related Work}

\subsection{6-Dof Grasp Detection}

With advancements in deep learning and the availability of large-scale grasp detection datasets, the performance of grasp detection models has seen substantial improvement. S4G \cite{qin2020s4g} extracts features from single-view point cloud and regress grasp poses and grasp quality scores. REGNet \cite{zhao2021regnet} builds a three-stage network to perform three tasks: sampling grasp points, generating grasp candidates, and optimizing grasp poses. Wang et al. \cite{wang2021graspness} propose GSnet, which samples seed points from the scene point cloud and aggregates features in the cylindrical area around them to generate grasping configurations. Tang et al. \cite{tang2024high} apply multi-radius cylindrical sampling to fuse local features and refine the distance between grasping candidate poses by upsampling. Dai et al. \cite{dai2023graspnerf} propose the GraspNeRF framework to detect 6-DoF grasp poses of transparent objects based on multiple RGB images. Ma et al. \cite{Ma2024generalized} use domain prior knowledge to enhance the generalization ability of the grasp detection network. In contrast to existing literature, our research delves into more effective methods for feature extraction, introducing hierarchical heatmaps to improve both the quality and efficiency of grasp detection.

\subsection{Efficient Grasp Detection}

The emergence of large-scale grasping datasets has established a robust data foundation for developing high-precision models. However, this advancement also brings increased training time, model size, and resource consumption. Nie et al. \cite{nie2024smaller} present a lightweight 4-Dof grasping detection network that enhances performance and generalization through knowledge distillation. However, suction grasping and 4-DOF grasping are influenced by the structures of objects and scenes, making it challenging to achieve high-quality grasping, especially for irregular objects. HGGD \cite{chen2023hggd} utilizes heatmap to focus on the graspable region and speed up grasp detection, but it still relies on the 3D PointNet \cite{qi2017pointnet} and feature fusion to further detect grasps. Wu et al. \cite{wu2024economic} have designed an economic grasp detection framework that effectively reduces training resource consumption while maintaining grasping performance. Building on the concept of generative synthesis, Wong et al. \cite{Wong2023fast} introduce the Fast GraspNeXt framework, which efficiently handles multiple tasks, including the detection of occluded objects and the generation of suction grasp heatmaps. Different from the above works, we design a lightweight end-to-end framework with region feature propagation, achieving real-time inference capabilities on edge devices.

\begin{figure*}[t]
\centering
    \includegraphics[width=0.97\linewidth]{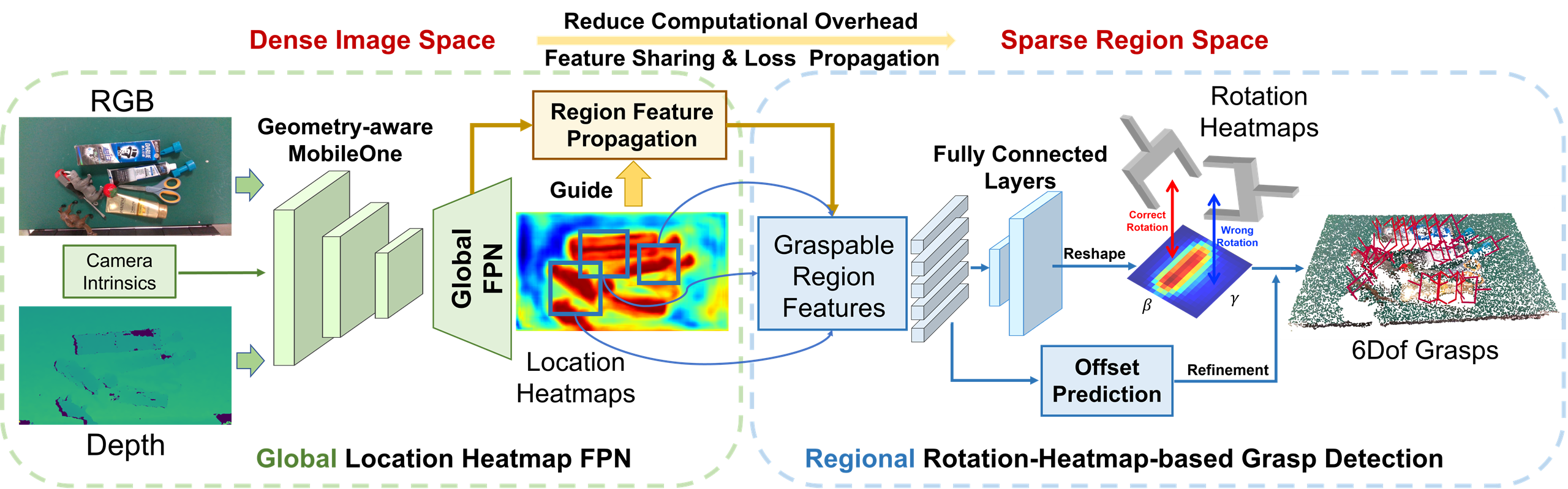}
    \caption{Overview of the proposed \textbf{Efficient End-to-End 6-Dof Grasp Detection Framework.}}
    \label{fig:framework} 
    \vspace{-0.5cm}
\end{figure*}

\section{Method}

Our efficient 6-DoF grasp detection framework consists of three primary components: the Global Location Heatmap FPN, Region Feature Propagation, and Regional Rotation-Heatmap-based Grasp Detection. As demonstrated in Fig. \ref{fig:framework}, the core innovation of our framework is its capability to effectively extract multi-scale features using a lightweight encoder and the Heatmap FPN, which are subsequently propagated to the Regional Rotation-Heatmap-based Grasp Detection stage, enabling precise predictions of grasp rotations. The overall methodology significantly reduces computational complexity while enhancing the quality and robustness of grasp detection.

\subsection{Problem Statement}

Our task is to efficiently generate a set of grasp poses $\mathbf{G}$ from a given single-view RGBD image $\boldsymbol{I} \in R^{ H \times W \times 4}$ and camera intrinsics. A 6-Dof grasp configuration $\boldsymbol{g}\in\mathbf{G}$ can be defined as: 
\begin{equation}
    \boldsymbol{g} = (x, y, z, \theta, \gamma, \beta, w),
\end{equation}
where $(x, y, z)$ is the 3-Dof location of the grasp pose and $(\theta,\gamma,\beta)\in [-\frac{\pi}{2}, \frac{\pi}{2}]$ represents the 3-Dof rotation of the grasp pose in the form of Euler angles in the gripper coordinate system. In addition, an extra parameter $w$ is predicted, which refers to the parallel-jaw width.

\subsection{Global Location Heatmap FPN}

For feature extraction in the dense image/3D space, differing from the heavy 3D encoder in previous methods \cite{fang2020graspnet, wang2021graspness, liu2022transgrasp}, we design our encoder in a more efficient fully-convolutional way. We draw inspiration from the RGB encoder MobileOne \cite{vasu2023mobileone} designed for mobile phones to create a lightweight encoder for RGBD feature extraction. However, the standard MobileOne is inadequate for processing RGBD images due to the loss of geometric information. To preserve crucial 3D geometry, we integrate camera intrinsics into the network by generating an additional positional mesh grid for each pixel in the RGBD image with the camera imaging model:
\begin{equation}
x = \frac{u - c_x}{1000\times f_x},\quad y = \frac{v - c_y}{1000\times f_y},
\end{equation}
where $(u,v)$ are the pixel coordinates and $(f_x, f_y, c_x, c_y)$ are the camera intrinsics. The encoder, named Geometry-aware MobileOne, takes the RGBD images and the mesh grid as a 6-channel input to extract semantic and geometry features. 

Moreover, recognizing the variability in grasp widths in real-world scenarios, we incorporate the widely used feature pyramid network (FPN) \cite{lin2017feature} to develop multi-scale scene features from the encoder’s multi-layer outputs. Building upon the effective location heatmap concept from \cite{chen2023hggd}, we design an additional heatmap prediction head after the FPN to leverage features and facilitate subsequent feature propagation. In summary, the Global Location Heatmap FPN, featuring the Geometry-aware MobileOne, effectively extracts multi-scale features, identifies graspable regions, and is optimized for deployment on edge devices.

\subsection{Region Feature Propagation}

Though more efficient than previous work, the two-stage framework introduced in \cite{chen2023hggd} exhibits a linearly increasing computational cost associated with the cropping of regions. Differing from HGGD, our approach propagates region-wise features from the FPN under the guidance of the grasp location heatmaps, thus reducing extra computation overhead via feature sharing. Instead of directly conducting feature pooling \cite{ren2016faster} on the feature maps, we introduce the grid sampling method proposed in \cite{jaderberg2015spatial} to obtain regional feature maps of multiple graspable regions with the same size $s$. First, we use Farthest Point Sampling \cite{qi2017pointnet++} to identify the region centers $c$ in the location heatmaps with high graspability values. Then, the local region grid indexes are generated according to the normalized depth values $d$ of the centers:
\begin{align}
    f_{reg}&= \text{gridsample}(f_{scene}, c + d \times \text{meshgrid}(-\frac{s}{2}, \frac{s}{2})),
\end{align}
which means the region feature maps are centered at $c$ with size $s\times s$. To preserve the features' continuity, we use bilinear interpolation in grid sampling. In contrast to a fixed cropping radius \cite{zhao2021regnet, chen2023hggd}, we introduce dynamic-scale region cropping by random sampling region size $s$ from a uniform distribution during model training to force our method to learn more robust feature representation. This process can also be applied during inference as a form of test-time augmentation to enhance performance. However, for efficiency, we typically conduct the grid sampling and the feature propagation process just once.

\subsection{Regional Rotation-Heatmap-based Grasp Detection}

As proved in pioneer work \cite{zhao2021regnet, fang2020graspnet, wang2021graspness, chen2023hggd}, anchor-based grasp detectors are efficient for grasp rotation prediction. Following \cite{chen2023hggd}, we construct a series of uniform rotation anchors for $(\gamma, \beta)$, whose Cartesian product forms a 2D rotation heatmap. Then, grasp detection can be framed as a semantic segmentation problem within the rotation heatmap. Instead of directly applying anchor-based classification, we enhance the grasp rotation precision by developing higher-resolution rotation heatmaps. In our implementation, to exploit the relationship between different rotations, we employ two fully connected layers to output the rotation heatmaps. Additionally, we incorporate a refinement module to produce grasp widths and additional center offsets, improving grasp detection quality.

\begin{table*}[t]
\scriptsize
\renewcommand{\arraystretch}{1.3}
\tabcolsep=0.15cm
\caption{Results on GraspNet Dataset, showing APs on the RealSense split}
\vspace{-0.4cm}
\begin{center}
\resizebox{0.8\textwidth}{!}{
\begin{threeparttable}
\begin{tabular}{c|c c c|c c c|c c c|c} 
\hline
\multirow{2}{*}{\textbf{Method}}& \multicolumn{3}{c|}{\textbf{Seen}} & \multicolumn{3}{c|}{\textbf{Similar}} & \multicolumn{3}{c|}{\textbf{Novel}} & {\textbf{Average}} \\\cline{2-11}
\multirow{2}{*}{} & $AP$ & $AP_{0.8}$ & $AP_{0.4}$ & $AP$ & $AP_{0.8}$ & $AP_{0.4}$ & $AP$ & $AP_{0.8}$ & $AP_{0.4}$ & m$AP$\\
\hline
GPD \cite{ten2017grasp} & 22.87 & 28.53 & 12.84 & 21.33 & 27.83 & 9.64 & 8.24 & 8.89 & 2.67 & 17.48\\
PointnetGPD \cite{liang2019pointnetgpd}  & 25.96 & 33.01 & 15.37 & 22.68 & 29.15 & 10.76 & 9.23 & 9.89 & 2.74 & 19.29\\ 
GraspNet \cite{fang2020graspnet}  & 27.56 & 33.43 & 16.95 & 26.11 & 34.18 & 14.23 & 10.55 & 11.25 & 3.98 & 21.41\\ 
TransGrasp \cite{liu2022transgrasp} & 39.81 & 47.54 & 36.42 & 29.32 & 34.80 & 25.19 & 13.83 & 17.11 & 7.67 & 27.65\\
HGGD$^\dagger$ \cite{chen2023hggd} & 58.35 & 66.54 & 55.96 & 47.93 & 56.91 & 41.86 & 22.10 & 27.37 & 14.31 & 42.79\\
SBGrasp \cite{ma2023towards}  & 58.95 & 68.18 & 54.88 & 52.97 
 & 63.24 & 46.99 & 22.63 & 28.53 & 12.00 & 44.85\\
GSNet \cite{wang2021graspness} & 65.70 & 76.25 & 61.08 & 53.75 & 65.04 & 45.97 & 23.98 & 29.93 & 14.05 & 47.81\\
\hline
E3GNet & \textbf{69.56} & \textbf{78.35} & \textbf{66.30} & \textbf{60.84} & \textbf{72.81} & \textbf{52.07} & \textbf{26.74} & \textbf{33.12} & \textbf{15.43} & \textbf{52.38}\\
\hline
\end{tabular}
$^*$The \textbf{best scores} are displayed in \textbf{bold}.\quad\quad$^\dagger$: Reimplemented with official codebase.\\
\end{threeparttable}}
\end{center}
\label{tab:graspnet}
\vspace{-1.0cm}
\end{table*}

\section{Experiments}

\subsection{Dataset, Metrics and Results}

We utilize the GraspNet-1Billion dataset \cite{fang2020graspnet} to train and evaluate our network, which comprises 100 real-world scenes, each containing RGBD images captured from 256 viewpoints. The test split contains 90 scenes, which are categorized into three sub-datasets: seen, similar, and novel, depending on the type of objects. We adopt the metric AP proposed by the GraspNet-1Billion dataset, which is the average of force-closure accuracy \cite{mahler2017dex}, AP$_{\mu}$, when the friction coefficient $\mu$ ranges from 0.2 to 1, with an interval of 0.2. For each scene, AP$_{\mu}$ represents the average grasp precision of the top 50 grasps after non-maximum suppression to evaluate grasp diversity.

We evaluate the overall performance of E3GNet against several grasp detection algorithms, all trained on the same dataset split. As indicated in Table \ref{tab:graspnet}, E3GNet demonstrates impressive results, achieving an average of 52.18 mAP across all the test scenes. Compared to other approaches, E3GNet surpasses the current state-of-the-art on all metrics, particularly excelling in the unseen categories, mainly benefiting from the well-designed feature extraction and propagation scheme.

\subsection{Model Efficiency Experiments}

In addition to the quality of grasp detection, the inference speed of the model significantly impacts real-world grasping execution, particularly on certain edge devices. To thoroughly evaluate model efficiency, we tested various models across two types of devices, totaling four different platforms. We selected the one-time inference time—from data input to grasp output—as our evaluation metric. As shown in Table \ref{tab:speed}, E3GNet demonstrates superior performance in network inference speed on all tested platforms, highlighting the effectiveness of our framework design and network optimization.

\begin{table}[t]
\small
\renewcommand{\arraystretch}{1.3}
\caption{Model inference efficiency on different device platforms}
\vspace{-0.65cm}
\begin{center}
\resizebox{\linewidth}{!}{
\begin{threeparttable}
\begin{tabular}{c|>{\centering\arraybackslash}p{1.1cm} >{\centering\arraybackslash}p{1.1cm} >{\centering\arraybackslash}p{1.1cm} >{\centering\arraybackslash}p{1.1cm}} 
\hline
\multirow{2}{*}{\textbf{Platform}} & \multicolumn{4}{c}{\textbf{Method Inference Time / ms}}\\
\cline{2-5}
& \textbf{GSNet} & \textbf{GraspNet} & \textbf{HGGD}  & \textbf{E3Gnet}\\
\hline
\multicolumn{5}{c}{\textbf{Desktop Devices}} \\
\hline
\multirow{1}{*}{Nvidia RTX 3090$^1$} & 104.1 & 85.2 & 24.3 & \textbf{5.9}\\
\multirow{1}{*}{Nvidia RTX 3060Ti$^2$} & 154.2 & 102.0 & 30.7 & \textbf{9.2}\\
\hline
\multicolumn{5}{c}{\textbf{Edge Devices}} \\
\hline
\multirow{1}{*}{Nvidia Jetson TX2} & Failed & 424.8 & 649.4 & \textbf{157.9}\\
\multirow{1}{*}{Nvidia Jetson Xavier NX} & Failed & 363.1 & 418.1 & \textbf{126.7}\\
\hline
\end{tabular}
\label{tab:speed}
$^*$All time are averaged by 200 trials and shown in miliseconds.\\
$^1$Intel 13900KF CPU, Nvidia RTX 3090 GPU and ubuntu 20.04.\\
$^2$AMD 5600X CPU, Nvidia RTX 3060Ti GPU and ubuntu 20.04.\\
\end{threeparttable}}
\vspace{-0.6cm}
\end{center}
\end{table}

\begin{table}[tb]
\small
\renewcommand{\arraystretch}{1.25}
\caption{Method component ablation, shwoing result on Realsense split and time tested on the Nvidia Jetson TX2 platform}
\vspace{-0.7cm}
\begin{center}
\resizebox{\linewidth}{!}{
\begin{threeparttable}
\begin{tabular}{c c c|c c c c} 
\hline
\multicolumn{3}{c|}{\textbf{Components}}&\multirow{2}{*}{\textbf{Seen}}&\multirow{2}{*}{\textbf{Similar}}&\multirow{2}{*}{\textbf{Novel}}&\multirow{2}{*}{\textbf{Time/ms}}\\
\emph{RFP} & \emph{GLH} & \emph{RRH} & & & &\\
\hline
& & & 58.35 & 47.93 & 22.10 & 649.4\\
\checkmark & & & 68.20 & 58.96 & 26.58 & 186.4\\
\checkmark & \checkmark & & 67.62 & 58.72 & 26.53 & 158.2\\
\checkmark & \checkmark & \checkmark & 69.56 & 60.84 & 26.74 & 157.9\\
\hline
\end{tabular}
\label{tab:ablation}
\emph{RFP}: \textbf{R}egion \textbf{F}eature \textbf{P}ropagation\\
\emph{GLH}: \textbf{G}lobal \textbf{L}ocation \textbf{H}eatmap FPN\\
\emph{RRH}: \textbf{R}egional \textbf{R}otation-\textbf{H}eatmap-based Grasp Detection\\
\end{threeparttable}}
\vspace{-1.0cm}
\end{center}
\end{table}

\subsection{Ablation Studies}

Ablation studies are conducted on the three primary components of E3GNet, with HGGD \cite{chen2023hggd} as the baseline. The results presented in Table \ref{tab:ablation} underscore the impact of these components. As anticipated, our novel framework, incorporating Region Feature Propagation, significantly improves model inference efficiency by minimizing redundant computations and largely enhancing overall performance. Transitioning to a more efficient lightweight Global Location Heatmap FPN yields faster inference speeds, albeit with a slight decrease in grasp detection quality. Additionally, Rotation-Heatmap-based Grasp Detection moderately improves grasp detection quality without incurring any extra computational costs.

\begin{figure}[t]
\centering
    \includegraphics[width=0.75\linewidth]{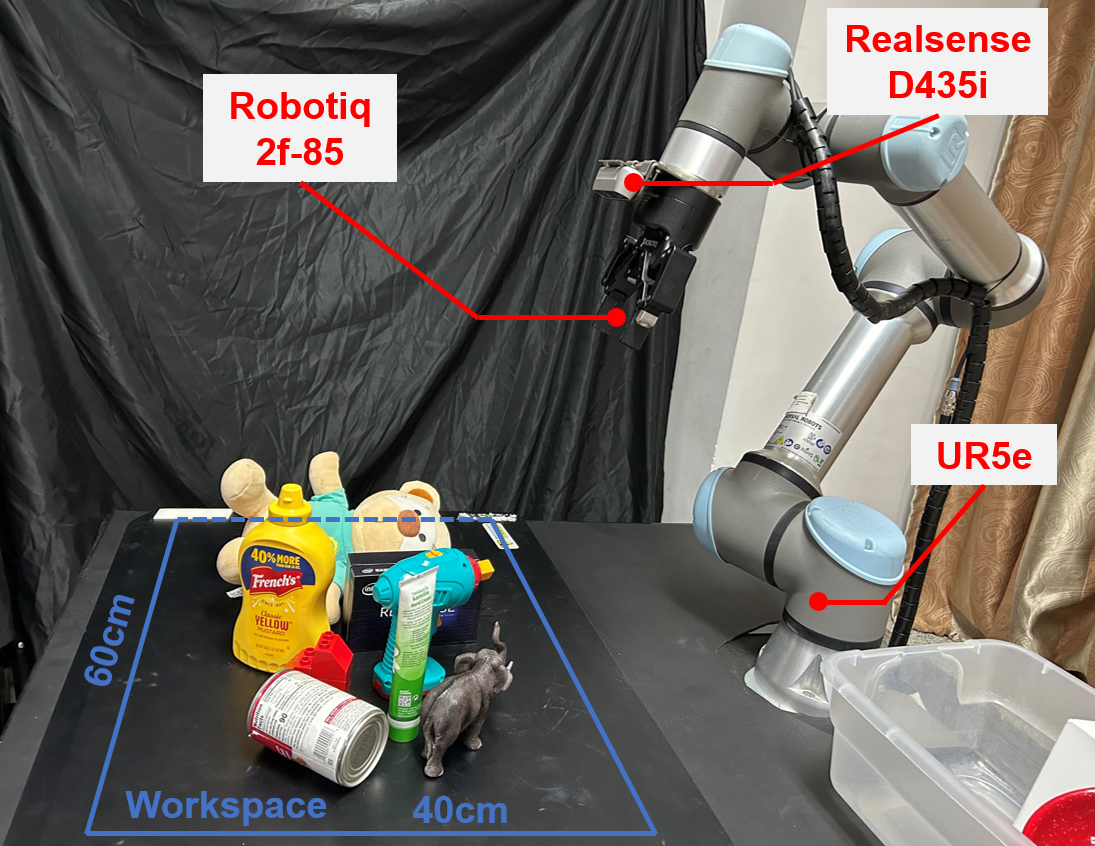}
    \vspace{-0.2cm}
    \caption{Real-world robot experiment settings.}
    \label{fig:realworld} 
    \vspace{-0.4cm}
\end{figure}

\begin{table}[t]
\small
\renewcommand{\arraystretch}{1.25}
\caption{Real-world clutter clearance results}
\vspace{-0.1cm}
\centering
\begin{tabular}{c c| c c} 
\hline
\textbf{Scene}& \textbf{Objects}& \textbf{E3GNet} & \textbf{HGGD}\\
\hline
1 & 7 & 7 / 8 & 7 / 7\\
2 & 8 & 8 / 8 & 8 / 11\\
3 & 8 & 8 / 9 & 7 / 12\\
4 & 8 & 8 / 9 & 8 / 11\\
5 & 8 & 8 / 8 & 8 / 8\\
6 & 9 & 9 / 9 & 9 / 10\\
\hline
\multicolumn{2}{c|}{\textbf{Success Rate}} & \textbf{94\%}(48 / 51) & 80\%(47 / 59) \\
\multicolumn{2}{c|}{\textbf{Completion Rate}} & \textbf{100\%}(6 / 6) & 83\%(5 / 6)\\
\hline
\end{tabular}
\label{tab:real}
\vspace{-0.6cm}
\end{table}

\subsection{Realworld Grasping Result}

To validate the effectiveness of E3GNet in real-world settings, we perform real-world robotic grasping experiments on a UR-5e robotic arm equipped with a Robotiq 2F-85 parallel-jaw gripper. A Realsense-D435i camera is mounted to capture ego-centric RGBD images. Our real-world experiment employs a collection of 22 objects with various shapes, generating 6 cluttered scenes composed of 7 to 9 randomly selected items in different poses. We focus on object grasping to clear these clutters. The robot conducts grasp detection and executes grasps until no grasps are identified or a maximum of 12 attempts are reached. Following \cite{fang2020graspnet, chen2023hggd}, we apply Success Rate (successful grasps / total attempts) and Completion Rate (cleared scenes / total scenes) as evaluation metrics.

Table \ref{tab:real} presents the performance of our method in real-world environments. Our approach achieves an impressive average grasp success rate of 94 \% and successfully clears all 6 scenes, demonstrating its ability to generate high-quality grasps. However, some failures were noted, particularly when the gripper collided with other objects in the clutters. Real-world demos and more implementation details can be found on the \href{https://kaiqinyang.github.io/E3GNet.github.io}{project page}.

\section{Conlcusion}

In this paper, we introduce an efficient end-to-end framework for 6-Dof grasp detection. By employing a hierarchical heatmap-based design, our method enables real-time detection of high-quality 6-Dof grasps and is suitable for deployment on edge devices. Our framework demonstrates state-of-the-art performance on the GraspNet-1Billion Dataset \cite{fang2020graspnet}. Additionally, quantitative experiments involving real-world robot grasping validate the effectiveness of our approach.

\bibliographystyle{IEEEtran}
\bibliography{reference}

\end{document}